\pdfoutput=1

\documentclass[11pt]{article}

\usepackage[final]{acl2021}

\usepackage{times}
\usepackage{latexsym}

\usepackage[T1]{fontenc}

\usepackage[utf8]{inputenc}

\usepackage{microtype}
\usepackage{amsmath,amsfonts,bm}

\usepackage{times}
\usepackage{latexsym}

\usepackage{url}
\usepackage{booktabs}
\usepackage{multirow}
\usepackage{array, caption, floatrow, makecell, booktabs}
\usepackage{wrapfig}
\usepackage{floatrow}
\usepackage{comment}
\usepackage{enumitem}
\usepackage{accents}
\usepackage{xspace,mfirstuc,tabulary}

\newcommand\blfootnote[1]{%
  \begingroup
  \renewcommand\thefootnote{}\footnote{#1}%
  \addtocounter{footnote}{-1}%
  \endgroup
}

\usepackage{todonotes}
\makeatletter
\newcommand*\iftodonotes{\if@todonotes@disabled\expandafter\@secondoftwo\else\expandafter\@firstoftwo\fi}  
\makeatother

\newfloatcommand{capbtabbox}{table}[][\FBwidth]

\newcommand*{\rom}[1]{\uppercase\expandafter{\romannumeral #1\relax}}

\title{Unsupervised Vision-and-Language Pre-training \\ Without Parallel Images and Captions}
{\centering
\author{Liunian Harold Li$^\dagger$, Haoxuan You$^{*\circ}$, Zhecan Wang$^{*\circ}$, Alireza Zareian$^\circ$, \\ \textbf{Shih-Fu Chang$^\circ$ \& Kai-Wei Chang$^\dagger$}\\

$^\dagger$University of California, Los Angeles\\$^\circ$Columbia University\\\texttt{liunian.harold.li@cs.ucla.edu,}\\
\texttt{\{hy2612,zw2627,az2407,sc250\}@columbia.edu,}\\\texttt{kwchang@cs.ucla.edu} \\
}
}
\newcommand{\nlvr}{NLVR$^2$\xspace}

\newcommand{\wvbfull}{Unsupervised VisualBERT\xspace}
\newcommand{\wvb}{U-VisualBERT\xspace}
\newcommand{\svb}{S-VisualBERT\xspace}
\newcommand{\wvbr}{U-VisualBERT$_\text{SBU}$\xspace}
\newcommand{\wvbnc}{U-VisualBERT$_\text{NC}$\xspace}

\newcommand{\wvbplus}{U-VisualBERT$_+$\xspace}
\newcommand{\hybird}{H-VisualBERT\xspace}

\newcommand{\wvbnt}{U-VisualBERT$_\text{NT}$\xspace}
\newcommand{\svbnt}{S-VisualBERT$_\text{NT}$\xspace}

\begin{document}

\maketitle

\begin{abstract}

Pre-trained contextual vision-and-language (V\&L) models have achieved impressive performance on various benchmarks. However, existing models require a large amount of parallel image-caption data for pre-training. Such data are costly to collect and require cumbersome curation. Inspired by unsupervised machine translation, we investigate if a strong V\&L representation model can be learned through unsupervised pre-training without image-caption corpora. In particular, we propose to conduct ``mask-and-predict'' pre-training on text-only and image-only corpora and introduce the object tags detected by an object recognition model as anchor points to bridge two modalities. We find that such a simple approach achieves performance close to a model pre-trained with aligned data, on four English V\&L benchmarks. Our work challenges the widely held notion that aligned data is necessary for V\&L pre-training, while significantly reducing the amount of supervision needed for V\&L models.

\end{abstract}

\section{Introduction}
\blfootnote{$^*$The two authors contributed equally.}

Pre-trained contextual vision-and-language (V\&L) models \citep{lu2019vilbert,tan2019lxmert,li2019visualbert,su2019vl,chen2020uniter} have achieved high performance on various V\&L tasks. However, different from contextual language models, such as BERT \citep{devlin2019bert}, which are trained on easily-accessible unannotated text corpora, existing V\&L models are still a step away from self-supervision. They require a massive amount of aligned text-image pairs for ``mask-and-predict'' pre-training. Such aligned data are costly to collect and hard to scale up. For example, the widely used MS-COCO dataset \citep{chen2015microsoft} requires extensive annotation from crowd workers.\footnote{Other datasets also require cumbersome curation. For example, while Conceptual Captions is crawled from the web, the authors report that from 5 billion images gathered over the Internet, only 3 million have paired high-quality captions after filtering \citep{sharma2018conceptual,changpinyo2021conceptual}.} 

In this paper, we explore \emph{unsupervised V\&L pre-training} with unaligned image and text corpora.\footnote{Following \citet{lample2018unsupervised} and \citet{feng2019unsupervised}, we use the term ``unsupervised'' to refer to pre-training with unaligned data, while ``supervised'' refers to pre-training with aligned text and images.} This research direction aligns with the theme of unsupervised and self-supervised learning that moves from heavily-annotated data to unannotated data, e.g. unsupervised machine translation \citep{lample2018unsupervised} and unsupervised image captioning \citep{feng2019unsupervised}.
Unsupervised V\&L pre-training is highly desirable as in many domains, aligned data is scarce (e.g. multimodal hate speech detection \citep{kiela2020hateful} and the medical domain \citep{li2020comparison}) and it is easier to collect unaligned text and images. In addition to its practical implication, our endeavour challenges the widely held notion that image-caption corpora is indispensable for pre-training \citep{lu2019vilbert} and brings valuable insight into the role that aligned data play in V\&L pre-training. 

\begin{figure*} [t]
\centering
\includegraphics[width=1.0\textwidth]{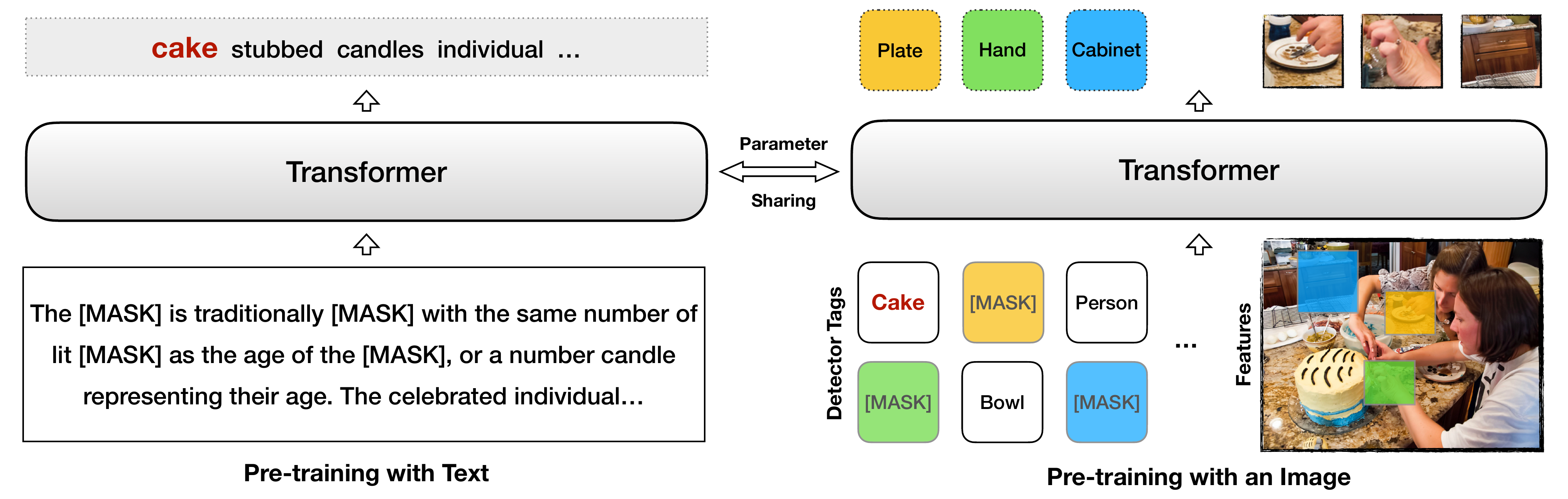}
\caption{An illustration of pre-training without aligned data. Given text, the model is trained to predict masked words; given an image, the model is trained to predict masked regions and detector tags. The semantic class ``cake'' appears in both the language modality and the visual modality and is linked through the detector tags. Note that we do not require a text segment with the word \textit{cake} to appear together with the image. Rather, we assume that as long as the text corpora are general enough, the word \textit{cake} will appear in the textual modality eventually. The model can thus learn V\&L representations from such weak supervision signals.
}
\label{fig:model}
\end{figure*}

We are inspired by works on multi-lingual contextual language models \citep{pires2019how}. If we treat an image as a set of regions and each region as a visual token \citep{dosovitskiy2020image}, V\&L models share a similar goal with multi-lingual models as they both learn shared representations across different domains.
Although a multi-lingual language model pre-trained on non-parallel corpora such as mBERT \citep{devlin2019mbert} cannot align or translate languages out-of-the-box, its representation spaces for different languages can be easily aligned with a linear probe \citep{conneau2020emerging}. This property suggests the existence of universal latent symmetries in the unaligned contextual embedding spaces and is believed to contribute to mBERT's cross-lingual transfer ability. Thus we hypothesize that strong V\&L representations can be similarly learned by ``mask-and-predict'' pre-training on unaligned language and vision data. 

We propose unsupervised V\&L pre-training with unaligned text and images (see an illustration in Figure \ref{fig:model}). 
Specifically, we take VisualBERT \citep{li2019visualbert} as a running example and apply unsupervised pre-training, resulting in \textbf{\wvbfull} (\wvb). The model takes the form of a single Transformer that can accept inputs from both modalities. During each step of pre-training, unlike the existing models that observe a batch of text-image pairs, our model observes either a batch of text segments or a batch of images. When provided with text, part of the text is masked and the model is trained to predict the masked words; when provided with an image, part of the image regions are masked and the model is trained to predict properties of the masked regions.

To further encourage cross-modal fusion, we leverage the tags from an object detector as ``anchor points'' \citep{li2020oscar}. For every object, we append its detected tag as a word to the visual input. The mask-and-predict objective is applied to the tags. For instance, for the image in Figure \ref{fig:model}, the model can observe ``\textit{cake}'' appears naturally as a word, a tag, and an image region. The direct typing of image regions and words can be learned and serves as a starting point for further alignment. The function of the detector tags resembles that of the ``overlapping vocabulary'' in multi-lingual language models, i.e., identical strings that appear in different languages with the same meanings (e.g., ``DNA'' appears in both English and French). As the ``overlapping vocabulary'' improves cross-lingual transfer \citep{wu2019beto}, we argue the detector tags can improve cross-modal grounding. 

We first conduct controlled experiments by pre-training on an English image-caption corpus without providing the alignment, following unsupervised machine translation and image captioning \citep{gu2019unpaired}. Results on four English V\&L benchmarks (VQA \citep{goyal2017making}, \nlvr \citep{suhr2018corpus}, Flickr30K Image Retrieval \citep{plummer2015flickr30k}, and RefCOCO+ \citep{yu2016modeling}) show that \wvb achieves comparable performance as models with access to text-image pairs (Section \ref{exp1}). 

Additionally, our approach is effective in practical settings, 1) when using independently collected images and captions and 2) when using images and general-domain text (BookCorpus \citep{zhu2015aligning}) without any captions (Section \ref{ablation:rq1}). Quantitative and qualitative analysis confirms the anchoring effect of the detector tags (Section \ref{ablation:rq2}). As a byproduct, we conduct preliminary experiments to show the promise of the approach in a semi-supervised setting, where a hybrid model pre-trained with both aligned and additional unaligned data surpasses a model pre-trained only on aligned data. (Section \ref{ablation:rq3}).
The above experiments demonstrate the wide applicability of our method. We will open-source the project under \url{https://github.com/uclanlp/visualbert}.

\section{Related Work}
\label{relatedwork}
\paragraph{Pre-trained V\&L Transformers}
Various V\&L models that are pre-trained with a ``mask-and-predict'' objective on aligned text-image data have been proposed \citep{lu2019vilbert,tan2019lxmert,li2019visualbert,su2019vl,chen2020uniter,li2020unicoder,zhou2020unified,huang2020pixel,yu2020ernie,gan2020large}. Two kinds of designs have been proposed. Two-stream models \citep{lu2019vilbert,tan2019lxmert,yu2020ernie} utilize separate Transformers \citep{vaswani2017attention} for each modality and a cross-modality module is adopted. Single-stream models \citep{li2019visualbert,su2019vl,chen2020uniter} directly input the text and visual embeddings into one single Transformer. They have been widely used by downstream tasks \citep{kiela2020hateful}. Probing tasks \citep{cao2020behind} confirm that they capture useful V\&L information after pre-training.

Two studies also try to incorporate ``tag'' information during pre-training. Oscar \citep{li2020oscar} adds detected tags as additional signals when pre-training with aligned data. We, however, do so for pre-training with unaligned data and show that the tags serve a more important role in unsupervised pre-training (Section \ref{exp:rq3}). VIVO \citep{hu2020vivo} targets novel object captioning. They use manually annotated image-tag data for pre-training and image-caption data for fine-tuning. We do not use manually annotated data and the tags are noisily generated by a detector.

\paragraph{Self-supervised Representation Learning}
Self-supervision involves creating supervision objectives from natural data, often by corrupting the input and training the model to reconstruct the input \citep{kolesnikov2019revisiting} or contrastive learning \citep{chen2020simple}. Self-supervised training on language \citep{peters2018deep,devlin2019bert} such as BERT has been proven useful for various NLP tasks \citep{liu2019linguistic}, while self-supervised visual representation learning has been centered around learning low-level visual features, in hope of enhancing the backbone CNN \citep{doersch2015unsupervised,pathak2016context,noroozi2016unsupervised,chen2020simple}. In this paper, we conduct V\&L pre-training by optimizing a reconstructive objective on unlabeled language-only and image-only data. Thus, our proposed model could be regarded as ``self-supervised''. Notably, our contextual visual representation is built on top of a pre-trained detector, operating at a level above local visual features.

\paragraph{Unsupervised Multi-lingual Language Model}
This work is inspired by multi-lingual representations trained without parallel corpora \citep{devlin2019mbert}. They are effective for cross-lingual transfer, which involves learning a model in one language and applying it to another with no additional training. Studies \citep{wu2019beto,conneau2020emerging} have confirmed several design choices that facilitate such transfer, e.g. shared parameters and overlapping vocabularies across languages, and we make similar design choices in \wvb (Section \ref{approach:our}).
We argue that multi-lingual representations bear resemblance to multi-modal representations as both seek to encode the alignment between two domains \citep{chen2020graph}.

\paragraph{Unsupervised Grounding Learning}
Prior works have explored learning grounding with weak or no supervision \citep{rohrbach2016grounding,xiao2017weakly,wang-etal-2020-maf}. Closest to this paper is unsupervised image captioning \citep{feng2019unsupervised,laina2019towards,gu2019unpaired}, which conducts image captioning with unpaired images and captions. Similar to this work, the detector tags serve as the anchor points for image captioning. However, unsupervised image captioning still requires captions, while our approach works with easy-to-collect general-domain text without any caption text (Section \ref{ablation:rq1}).

\section{Approach}
We first take Supervised VisualBERT (\svb) as an example and illustrate how a typical V\&L model is pre-trained with aligned data. Then we introduce unsupervised V\&L pre-training, and the resulting model \wvbfull (\wvb).

\label{approach}

\subsection{Background}
\label{approach:background}
As mentioned in Section \ref{relatedwork}, there are several V\&L representation learning methods based on BERT. We take Supervised VisualBERT (\svb) as an example, which will also be used as a baseline in the experiments. \svb is modified from the original VisualBERT \citep{li2019visualbert} and augmented with the visual objectives from LXMERT \citep{tan2019lxmert} and detector tags similar to Oscar \citep{li2020oscar} (discussed in detail in Section \ref{approach:our}).

Every input to \svb contains a text segment $T$ and an image $I$. The text and the image are first mapped into embedding vectors respectively. Text embeddings $\bm{T}$ is a matrix in which each column vector represents the embedding of a subword in the text sequence, i.e. $ \bm{T} = [\bm{w}_{1:n}]$. Following BERT, each subword embedding $\bm{w}_i$ is the sum of its token, position, and segment embedding. Image embeddings $\bm{I}$ include both the image region embeddings $\bm{r}_{1:m}$ and the detector tag embeddings $\bm{d}_{1:l}$ (see Section \ref{approach:our} for details). Each region embedding $\bm{r}_i$ is the sum of a visual feature vector from the detector and a spatial box coordinate embedding \citep{tan2019lxmert}. The text and visual embeddings are then passed through a Transformer to built contextual representations.

The model is pre-trained with a mask-and-predict objective. Given a text-image pair $[T,I]$ from the aligned dataset $D$, we randomly mask out some words $w_i$, some regions $r_j$, and some tags $d_k$ to obtain masked $[\tilde{T},\tilde{I}]$. The model is trained to predict the masked words, the properties of the masked regions, and the masked tags given $[\tilde{T},\tilde{I}]$. The pre-training objective can be summarized as:

\begin{equation*}
\min_\theta \sum_{[T,I] \in D} L_{T+I+M}\left(f_\theta( [\tilde{T},\tilde{I}]), [T,I]\right).
\label{eq1}
\end{equation*}
$f_\theta$ represents the embedding layer and the multi-layer Transformer. $L_{T+I+M}$ is the sum of 1) the masked language model loss $L_T$, 2) the image reconstruction loss $L_I$, and 3) an ``text-image match'' objective $L_{M}$. Specifically, $L_I$ includes a \textit{tag reconstruction} loss $L_I^{tag}$ (more details in Section \ref{approach:our}) and the two visual losses as in LXMERT \citep{tan2019lxmert}: the \textit{region feature regression} loss $L_I^{ref}$, which forces the model to regress to the visual vector, and the \textit{noisy label classification} loss $L_I^{cls}$, which predicts the detected labels of masked objects with the cross-entropy loss. With a probability of 0.5, we provide the model with a mismatched text-image pair instead of a matched pair, and $L_{M}$ asks the model to predict whether the image matches the text. After the model is pre-trained, it can be fine-tuned for V\&L tasks similar to how BERT is fine-tuned for NLP tasks.

\subsection{Unsupervised Pre-training}
\label{approach:our}
We introduce the two core design choices of unsupervised pre-training: mask-and-predict pre-training with unaligned data and the detector tags.

\paragraph{Mask-and-Predict Pre-training with Unaligned Data} 
We assume access to a text corpus $D_T$ and an image corpus $D_I$ for pre-training.
During every pre-training step, we randomly sample either a batch of text from $D_T$ or a batch of images from $D_I$. No alignment between text and images is provided to the model.
When pre-training with a text segment $T$, the model is trained to reconstruct $T$ given the masked $\tilde{T}$.\footnote{We adopt the next sentence prediction task in BERT when long documents are available.}  When pre-training with an image $I$, the model is trained to reconstruct $I$ given the masked $\tilde{I}$. A single Transformer is used throughout two modalities (i.e. $\theta$ shared across modalities). The pre-training objective can be summarized as:

\begin{equation*}
    \min_\theta  \sum_{T \in D_T} L_T(f_\theta (\tilde{T}), T) + \sum_{I \in D_I} L_I(f_\theta (\tilde{I}), I).
\end{equation*}
After pre-training, the model is fine-tuned on downstream tasks just as its supervised counterpart, with the input being a text-image pair.

\paragraph{Detector Tags} 
While mask-and-predict pre-training with unaligned data in itself achieves non-trivial performance (Section \ref{exp3}), we find it beneficial to provide noisy alignment signals in the form of the detector tags. When modeling an image $I$, for each region detected, we append the tag outputted by the object detector to the input. The detector \citep{ren2015faster} is pre-trained on a general object detection dataset \citep{krishna2017visual,anderson2018bottom} and the tags are essentially a bag of words that provide some noisy grounding signals to the model. During pre-training, we apply the mask-and-predict objective to the tags, which further encourages grounding.

We process the detector tags as a subword sequence $d_{1:l}$ with spatial coordinates.\footnote{Each tag corresponds to a region. A tag could be split into multiple subwords, so the total length of the tag subword sequence $l$ is equal to or larger than the number of regions $m$.} Every tag subword is embedded as the sum of its token embedding and a spatial coordinate embedding. The token embedding is the same as the token embedding used in text modeling, while the spatial coordinate embedding is the same as the coordinate embedding of the corresponding region. The coordinate embedding allows the model to distinguish tags from different regions.\footnote{This design differs from that of Oscar \citep{li2020oscar}. Oscar does not add the coordinate embeddings to tags to encourage the fusion of tag and visual representations.} With the detector tags added, the image $I$ is embedded as a sequence of image region features $\bm{r}_{1:m}$ followed by a sequence of detector tag embeddings $\bm{d}_{1:l}$, i.e. $\bm{I} = [\bm{r}_{1:m};\bm{d}_{1:l}]$. The tags are added during both pre-training and fine-tuning.
Further, during pre-training, certain tag subwords are masked and the \textit{tag reconstruction} loss $L_I^{tag}$ supervises the model to predict the masked tags. The tags are predicted just as masked subwords are predicted in text modeling. The prediction softmax layer is shared between the tag and text subwords.

The parameters involved in modeling tags include the token embedding, the coordinate embedding, and the subword softmax embedding. These embedding parameters are shared across modalities and encourage the model to project text, visual, and tag representations into the same space (see Section \ref{visualizaion} for an example). This resembles the design in multi-lingual language models, which use shared BPE embeddings and softmax weights across languages \citep{wu2019beto}.

\section{Experiment}
\label{exp1}

As the domain and quality of data may affect the model performance, the conventional practice in unsupervised learning is to use aligned corpora without providing alignments, allowing for controlled comparison with a supervised model. For example, unsupervised machine translation creates unaligned corpora by splitting up parallel corpora \citep{lample2018unsupervised} while unsupervised image captioning \citep{gu2019unpaired} create unaligned corpus by shuffling images and captions from MSCOCO \citep{chen2015microsoft}. Following prior work, we first conduct experiments by using Conceptual Captions (CC) \citep{sharma2018conceptual} as the source of images and text for both the supervised and unsupervised model. Later in Section \ref{ablation:rq1}, we show that our method is effective when the images and captions are collected independently and when no caption text is used. 

\begin{table*}[h]
\small
\caption{Evaluation results on four V\&L benchmarks. Our unsupervised model trained with unaligned data (\wvb) achieves close performance with a supervised model trained with aligned data (\svb). \wvb also rivals with several supervised models such as ViLBERT on most metrics.}
\label{table:rq1}
\begin{center}
\resizebox{\linewidth}{!}{
\begin{tabular}{
l@{\hspace{10pt}} | 
l@{\hspace{5pt}}|l@{\hspace{5pt}}l@{\hspace{10pt}} | 
l@{\hspace{10pt}}|
l@{\hspace{5pt}}l@{\hspace{10pt}}|
l@{\hspace{5pt}}l@{\hspace{5pt}}l@{\hspace{10pt}}|
l@{\hspace{5pt}}l@{\hspace{5pt}}l@{}}
\toprule

\multirow{2}{*}{Model} 
&  \multicolumn{1}{c|}{Aligned}  & \multicolumn{2}{c|}{Unaligned}
& \multicolumn{1}{c|}{VQA} 
&\multicolumn{2}{c|}{\nlvr}
&\multicolumn{3}{c|}{Flickr30K}
&\multicolumn{3}{c}{RefCOCO+}\\

 &  & Image & Text & Test-Dev  & Dev & Test-P & R@1 & R@5 & R@10 & Dev & TestA & TestB \\
\midrule

Pre-BERT & - & - & - & 70.22
& 54.1 & 54.8 
& 48.60 & 77.70 & 85.20 
& 65.33 & 71.62 & 56.02  \\

ViLBERT  & 3M & 0 & 0
& 70.55  
& - & - 
& 58.78 & 85.60 & 91.42 
& 72.34 & 78.52 & 62.61\\

VL-BERT & 3M & 0 & $\sim$50M
& 71.16  
& - & - 
& - & - & - 
& 71.60 & 77.72 & 60.99 \\

UNITER$_\text{cc}$ & 3M & 0 & 0
& \textbf{71.22 }
& - & -  
& - & - & - 
&  72.49 & 79.36 & 63.65\\

\svb
&3M  & 0 & 2.5M
& 70.87\textsubscript{$\pm$.02} 
& \textbf{73.44}\textsubscript{$\pm$.51}  & \textbf{73.93}\textsubscript{$\pm$.51}   
& \textbf{61.19}\textsubscript{$\pm$.06}  & \textbf{86.32}\textsubscript{$\pm$.12}  & \textbf{91.90}\textsubscript{$\pm$.02}  
&\textbf{73.65}\textsubscript{$\pm$.11}  & \textbf{79.48}\textsubscript{$\pm$.36}  & \textbf{64.49}\textsubscript{$\pm$.22}   \\

\midrule

Base  & 0 & 0 & 0
& 69.26 
& 68.40 & 68.65  
& 42.86 & 73.62 & 83.28  
& 70.66 &  77.06  & 61.43 \\

\wvb & 
0 & 3M & 5.5M 
& \textbf{70.74}\textsubscript{$\pm$.06}  
& \textbf{71.74}\textsubscript{$\pm$.24} & \textbf{71.02}\textsubscript{$\pm$.47}  
& \textbf{55.37}\textsubscript{$\pm$.49} & \textbf{82.93}\textsubscript{$\pm$.07} & \textbf{89.84}\textsubscript{$\pm$.21}
&\textbf{72.42}\textsubscript{$\pm$.06} & \textbf{79.11}\textsubscript{$\pm$.08}& \textbf{64.19}\textsubscript{$\pm$.54}\\

\bottomrule
\end{tabular}
}
\end{center}
\end{table*}

\paragraph{\wvb} The model is pre-trained with shuffled captions and images. At each training step, we sample either a batch of images or a batch of text. Following VL-BERT \citep{su2019vl}, we find it beneficial to include BookCorpus \citep{zhu2015aligning}, a general-domain text corpus, during pre-training. In sum, \wvb is trained on 3M images from CC, 3M captions from CC, and 2.5M text segments from BookCorpus\footnote{Our version of BookCorpus contains around 5M text segments with 64 words per segment. For computational reasons, we downsample the dataset such that during each epoch, the model observes only half of the text segments from BookCorpus. This downsampling is also done for the other VisualBERT variants.}.

\paragraph{\svb} We introduce a Supervised VisualBERT (\svb) trained with aligned data as introduced in Section \ref{approach:background}. \svb is pre-trained on 3M caption-image pairs from CC and 2.5M text segments from BookCorpus.

\paragraph{Compared Models} Additionally, we list the performance of a \textbf{Base} VisualBERT that is initialized from BERT and does not undergo further pre-training. Previously reported supervised models that are trained on CC are also listed, including \textbf{ViLBERT}, \textbf{VL-BERT}, and \textbf{UNITER}. For UNITER, we include the version that is trained only on CC (UNITER$_\text{cc}$)\footnote{The results are from Appendix A.6 of \citet{chen2020uniter}.}. Although their network architectures differ from ours and cannot be directly compared, they jointly paint the picture of the performance we should expect by pre-training on CC. Models developed before BERT are listed as \textbf{Pre-BERT} (\citet{gao2019dynamic} for VQA, \citet{suhr2018corpus} for \nlvr, \citet{lee2018stacked} for Flickr30K, and \citet{yu2018mattnet} for RefCOCO+).

\paragraph{Setup} For all the VisualBERT variants introduced in the paper, we initialize them from BERT$_{base}$ and pre-train for 10 epochs on their respective pre-training datasets with a batch size of 144. All models can be trained within 3 days on 4 V100s each with 16GB of memory. We use the Adam optimizer \citep{kingma2014adam} with a linear-decayed learning-rate schedule \citep{devlin2019bert} and a peak learning rate at $6\times10^{-5}$. We conduct evaluations by fine-tuning on four downstream tasks: Visual Question Answering (VQA 2.0) \citep{goyal2017making}, Natural Language for Visual Reasoning (\nlvr) \citep{suhr2018corpus}, Image Retrieval (Flickr 30K) \citep{plummer2015flickr30k}, and Referring Expression (RefCOCO+) \citep{yu2016modeling}. We use a Faster R-CNN pre-trained on the Visual Genome dataset to extract region features \citep{anderson2018bottom}. For each task, we follow the recommended setting in previous works. For details, please refer to the appendix.

\begin{table*}[h]
\caption{Unsupervised pre-training is applicable when images and captions are collected independently (\wvbr) or when no caption text is provided (\wvbnc).}
\label{table:rq2}
\begin{center}
\resizebox{\linewidth}{!}{
\begin{tabular}{l@{\hskip9pt} | 
c@{\hskip9pt} c@{\hskip9pt}|c@{\hskip9pt}
c@{\hskip9pt}c@{\hskip9pt}c@{\hskip9pt}
c@{\hskip9pt}c@{\hskip9pt}c@{\hskip9pt} c@{\hskip9pt}
c@{\hskip9pt}c@{\hskip9pt}c@{\hskip9pt}c@{\hskip9pt}c@{\hskip9pt}c}
\toprule

\multirow{2}{*}{Model} & \multicolumn{2}{c|}{Text}  &
\multicolumn{1}{c}{VQA} & & \multicolumn{2}{c}{\nlvr} & & \multicolumn{3}{c}{Flickr30K} & & \multicolumn{3}{c}{RefCOCO+} \\

 & Caption & General & Test-Dev  & & Dev & Test-P & & R@1 & R@5 & R@10 & & Dev & TestA & TestB \\

\midrule
Base  & - & -
& 69.26 &
& 68.40 & 68.65  &
& 42.86 & 73.62 & 83.28  &
& 70.66 &  77.06  & 61.43 \\

\wvb 
&  CC & BC
& \textbf{70.74} &
& 71.74 & 71.02  &
& 55.37 & \textbf{82.93} & 89.84  &
& 72.42 & 79.11 & 64.19\\

\wvbr
& SBU & BC
& 70.70  &
& \textbf{71.97} & \textbf{72.11}  &
& \textbf{56.12} & 82.82 & \textbf{90.12} &
& \textbf{73.05} & \textbf{79.48} & 64.19 \\

\wvbnc
& - & BC 
& 70.47 & 
& 71.47 & 71.19  &
& 54.36 & 82.22 & 89.24  &
& 72.96 & 79.30 & \textbf{64.25}  \\

\bottomrule
\end{tabular}
}
\end{center}
\end{table*}

\paragraph{Results}
Table \ref{table:rq1} summarizes the results. For each model, we list the type and amount of data used during pre-training.\footnote{For models initialized from BERT, we do not count the BERT pre-training data. VL-BERT uses both BookCorpus and Wikipedia during V\&L pre-training. We estimate that the two corpora roughly have 5OM segments with 64 words per segment. With a different pre-processing style (e.g. longer segments), the number of segments may change.} To control for randomness, we report the means and standard deviations of \wvb and \svb across three runs.

\wvb outperforms the Base model on all benchmarks, while only lagging behind \svb slightly on VQA, \nlvr, and RefCOCO+. \wvb even surpasses or rivals with some supervised models (e.g., ViLBERT on VQA and RefCOCO+, VL-BERT on RefCOCO+, and UNITER$_\text{cc}$ on RefCOCO+). This shows that a model through unsupervised pre-training can perform comparably with supervised models. 

On Flickr30K Image Retrieval, the difference between \wvb and \svb is more evident. The task focuses on identifying if an image and a text segment are coherent. \svb is provided with explicit signals for such a task with the ``text-image match'' objective $L_M$ during pre-training (Section \ref{approach:background}). While \wvb is not provided with such explicit signals, it still performs better than the Base model. Further, if we were to remove the explicit signal (i.e. the ``text-image match'' objective) when pre-training on aligned data, \svb without $L_M$ achieves only 57.98 on R@1, much closer to \wvb 

\begin{table*}[h]
\small
\caption{Detector tags show a larger impact in the unsupervised setting (\wvbnt vs. \wvb) than in the supervised setting (\svbnt vs. \svb). Semi-supervised pre-training (\hybird) shows marginal improvement over supervised pre-training (\svb).}
\label{table:rq3}
\begin{center}
\resizebox{\linewidth}{!}{
\begin{tabular}{l@{\hskip10pt} | 
l@{\hspace{10pt}}|
l@{\hspace{5pt}}l@{\hspace{10pt}}|
l@{\hspace{5pt}}l@{\hspace{5pt}}l@{\hspace{10pt}}|
l@{\hspace{5pt}}l@{\hspace{5pt}}l@{}}
\toprule

\multirow{2}{*}{Model}  
& \multicolumn{1}{c|}{VQA} 
& \multicolumn{2}{c|}{\nlvr} 
& \multicolumn{3}{c|}{Flickr30K} 
& \multicolumn{3}{c}{RefCOCO+} \\

 & Test-Dev 
 & Dev & Test-P 
 & R@1 & R@5 & R@10 
 & Dev & TestA & TestB \\

\midrule

Base$_\text{NT}$ 
& 69.06 
& 51.98 & 52.73  
& 48.40 & 78.20 & 87.18  
& 70.15 & 76.91 & 61.72  \\

\midrule
\wvbnt
& 69.87 
& 67.90 & 68.92 
& 50.56 & 80.22 & 88.32 
& 71.94 & 77.79 & 62.38 \\

\wvb 
& 70.74 
& 71.74 & 71.02 
& 55.37 & 82.93 & 89.84 
& 72.42 & 79.11 & 64.19\\

\midrule

\svbnt
& 70.49  
& 72.56 & 73.53 
& 60.26 & 85.58 & 91.64 
& 72.70 & 77.93 & 62.99 \\

\svb
& 70.87 
& 73.44 & 73.93  
& 61.19 & 86.32 & 91.90  
& 73.65 & 79.48 & 64.49  \\

\midrule
\hybird
& 71.05\textsubscript{$\pm$.02} 
& 73.80\textsubscript{$\pm$.26}  & 74.82\textsubscript{$\pm$.25}  
& 60.28\textsubscript{$\pm$.60} & 86.30\textsubscript{$\pm$.35} & 92.06\textsubscript{$\pm$.28}  
& 74.01\textsubscript{$\pm$.25} & 80.18\textsubscript{$\pm$.23} & 64.89\textsubscript{$\pm$.24} \\

\bottomrule
\end{tabular}
}
\end{center}
\end{table*}

\section{Analysis}
In this section, we analyze the effect of the text data and the role of the detector tags.

\subsection{The Effect of Text Data}
\label{ablation:rq1}
The assumption behind unsupervised pre-training is that the detector tags should appear both in the images and text corpus, serving as the grounding anchor points. When the images and captions come from the same corpus, such an assumption clearly holds, and unsupervised pre-training works well (Section \ref{exp1}). However, we are curious if such an assumption still holds 1) if images and captions come from independently collected corpora (\wvbr) and 2) if no caption text but general-domain text is provided (\wvbnc). 

The latter setting bears great practical value. Conceptually, collecting caption-style text could be as hard as collecting image-caption data as images and captions seldom appear separately. It is desirable to explore training V\&L representations without caption-style text. Thus we experiment pre-training with general-domain text, which could be easier to collect.

\paragraph{\wvbr} We use 3M images from CC and 1M captions from SBU captions \citep{ordonez2011im2text}. To compensate for the different amounts of text between CC and SBU, we upsample the BookCorpus so that the amount of text data used by \wvbr is roughly the same as \wvb.

\paragraph{\wvbnc} The model is trained on images from CC and text from BookCorpus, a general-domain corpus.

\paragraph{Results} Unsupervised pre-training is effective in both scenarios (Table \ref{table:rq1}). When pre-training images and text are collected independently, \wvbr achieves similar performance as \wvb, with the latter higher on VQA, and the former higher on the other three tasks. 

When no caption text is used, the performance on \nlvr and RefCOCO+ remains unaffected while the performance on VQA and Flickr30K drops slightly, potentially because the language style of VQA and Flickr30K is similar to captions, benefiting \wvb. Such results are not surprising. In general-domain corpora like Wikipedia, grounded words take up a decent portion ($>$25$\%$) \citep{tan2020vokenization}. Thus the tags appear in pre-training text corpora with a non-trivial frequency and \wvbnc learns from such signals.
The above results suggest the applicability of unsupervised pre-training to many language-only and image-only datasets, which are easier to collect than image-caption datasets \citep{trinh2018simple,sun2017revisiting}.

\subsection{The Detector Tags as Anchor Points}
\label{exp3}
\label{exp:rq3}
\label{ablation:rq2}

We study the effect of the detector tags in unsupervised and supervised pre-training, respectively.

\paragraph{\textbf{W-VisualBERT$_{\text{NT}}$}}
\wvbnt observes no tags and only dense region features for image embeddings during pre-training and fine-tuning. For comparison, a base model  without tags is introduced (Base$_\text{NT}$), which is initialized from BERT and does undergo further pre-training.

\paragraph{\textbf{S-VisualBERT$_{\text{NT}}$}}  To study the effect of the detector tags when aligned data are present, we introduce \svbnt which is trained on aligned data but observes no tags for image embeddings.

\paragraph{Result}
We first find that even without tags, unsupervised pre-training benefits downstream tasks (Table \ref{table:rq3}). \wvbnt outperforms Base$_\text{NT}$ on all metrics with a large margin. We attribute this to the (unaligned) contextual V\&L representation learned through pre-training. This bears resemblance to the observation in multi-lingual language models that the shared vocabulary across languages (i.e. anchor points) is not necessary for cross-lingual transfer \citep{conneau2020emerging}.

Further, while the detector tags are beneficial for both supervised and unsupervised pre-training, the performance improvement is more evident for the latter. For example, performance difference on VQA between \wvb and \wvbnt is 0.95 (70.82 vs. 69.87) while the difference between \svb and \svbnt is 0.41 (70.90 vs. 70.49). The results are expected. When aligned data are present, object tags serve as additional signals while in unsupervised pre-training, they serve as the only source from which grounding is learned.

\begin{figure*} [t]
\centering
\includegraphics[width=1.0\textwidth]{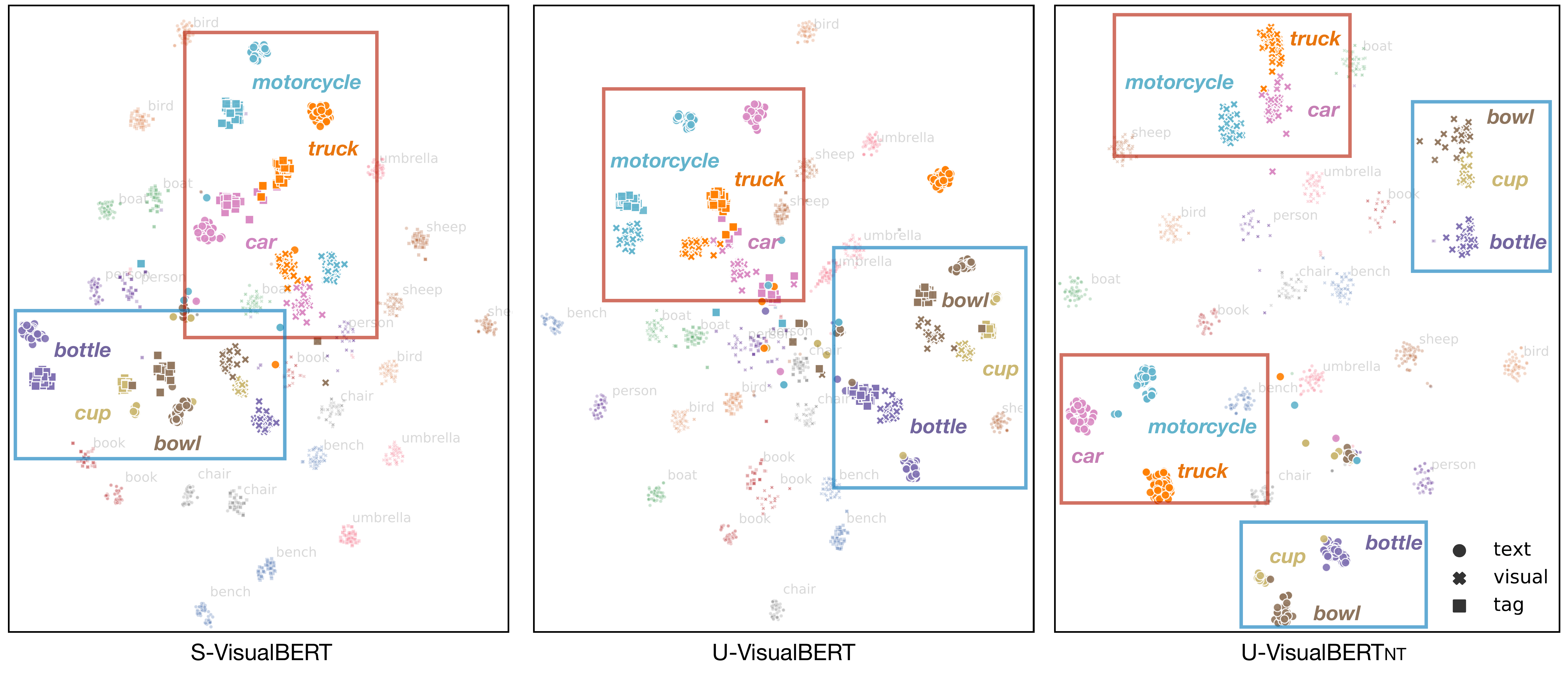}
\caption{Visualization of the contextual representations of \svb, \wvb, and \wvbnt. The tags help to fuse text and visual representations for \svb and \wvb. In \wvbnt, common structures emerge in the text and visual representation spaces even though they are not aligned.}
\label{fig:visualization}
\end{figure*}

\paragraph{Visualization}
\label{visualizaion}
To gain a direct sense of how the detector tags help bridge the modalities, we visualize the contextual representation spaces of \svb, \wvb, and \wvbnt in Figure \ref{fig:visualization}. For each of the most frequent 15 object classes in the COCO dataset \citep{chen2015microsoft}, we randomly sample at most 50 instances and take the last-layer contextual representations of the words, the objects, and the tags (when available) and visualize them with t-SNE \citep{maaten2008visualizing}. We highlight the representations of six selected classes.

Though trained without aligned data, \wvb can group text, tag, and visual representations by their semantic classes. Similar phenomena can be observed in \svb. \wvbnt, lacking any signal to align the two spaces, does not show signs of such behaviour. In \wvbnt, text and visual representations are almost completely separated (e.g., the two \textit{disjoint red} rectangles in the figure on the right). However, some common structures emerge in both modalities. For instance, representations for ``car'', ``truck'', and ``motorcycle'', the three semantically-related classes, are close to each other, in both the textual and visual modality (the red rectangles); representations for ``cup'', ``bottle'', and ``bowl'' are close (the blue rectangles). This also holds for the other two models and resembles what is observed in \citet{li2020oscar} and \citet{ilharco2020probing}.

\section{Semi-Supervised Pre-Training}
\label{exp:rq2}
\label{exp2}
\label{ablation:rq3}

Unsupervised pre-training in itself has great practical and research value in many domains where aligned data is scarce. As a byproduct, we wonder if the approach could find its use in a semi-supervised setting, where we pre-train a model with both aligned data and unaligned data.

\paragraph{\hybird} We introduce a \textit{hybrid} model that is trained on the 3M aligned data from Conceptual Captions (CC) and additional unaligned 1.7M images from Open Images (OI) \citep{kuznetsova2020open}. When a training sample comes from CC, we provide the model with a text-image pair, and when the training sample comes from OI, we provide only the image. We do not use any manually annotated visual labels provided in OI. 

\paragraph{Result}
We control for randomness by running \hybird for three times and report the means and stand deviations. We observe that \hybird brings consistent improvement upon \svb on most tasks (Table \ref{table:rq3}) except Flickr30K\footnote{On Flickr30K, the performance between \hybird and \svb is similar, potentially because the ``image-text match'' objective is the dominant contributor and additional image-only data during pre-training have limited benefit (Section \ref{exp1}).}. This preliminary result is promising as the dataset scale in this experiment is relatively small (million-scale). Meanwhile, unannotated data generally could not improve upon a model trained with annotated data significantly, unless drastically scaled up \citep{he2020momentum}. We leave large-scale experiments to future work.

\section{Conclusion}
In this paper, we explore unsupervised pre-training with unaligned data. We conduct mask-and-predict pre-training on textual data and visual data and the detector tags are used as anchor points to bridge the two modalities. Experiments show that unsupervised pre-training can achieve performance similar to supervised pre-training. 

\section*{Ethical Considerations}
One caveat of the proposed method is that data collected from the web may contain biases \citep{zhao2017men}, toxic contents \citep{schmidt2017survey}, and other ethical issues. This problem is common to ML models and we stress that de-biasing \citep{zhao2019gender} and a rigorous examination are needed before deploying the system.

\section*{Acknowledgement}
We would like to thank Hao Tan, members of UCLA NLP, and members of UCLA PlusLab for their helpful comments. We also thank the reviewers for the valuable reviews. This work was supported in part by DARPA MCS program under Cooperative Agreement N66001-19-2-4032. The views expressed are those of the authors and do not reflect the official policy or position of the Department of Defense or the U.S. Government.

\bibliography{iclr2019_conference,nlp,preprint,ref,newref,acl2021}
\bibliographystyle{acl_natbib}

\appendix

\section{Fine-Tuning on Downstream Tasks}

We describe the details of fine-tuning on the four downstream tasks: Visual Question Answering (VQA 2.0) \citep{goyal2017making}, Natural Language for Visual Reasoning (\nlvr) \citep{suhr2018corpus}, Image Retrieval (Flickr 30K) \citep{plummer2015flickr30k}, and Referring Expression (RefCOCO+) \citep{yu2016modeling}.

\paragraph{VQA} Given an image and a question, the task is to correctly answer the question. We use the VQA 2.0 and use the Karpathy split for training and validation \citep{karpathy2015deep}. We fine-tune with a binary cross-entropy loss. The model is trained with a batch size of 32 and a peak learning rate of $5\times10^{-5}$ over 8 epochs.

\paragraph{\nlvr} \nlvr involves determining whether a natural language caption is true about a pair of images. While more sophisticated fine-tuning strategy exists \citep{chen2020uniter}, we follow LXMERT \citep{tan2019lxmert} to pair the caption with each image, concatenate the ``[CLS]'' representation of the two pairs, and build a classifier on top. We find it beneficial to conduct a moderate amount of ``task-specific pre-training'' where we use the data from the dataset to conduct mask-and-predict pre-training as suggested by VisualBERT \citep{li2019visualbert}. We conduct task-specific pre-training for at most 5 epochs and fine-tune from the epoch with the best validation LM loss. Fine-tuning is conducted for 8 epochs with a batch size of 32 and a peak learning rate of $2\times10^{-5}$.

\paragraph{Flickr30K}  The task of image retrieval involves finding the corresponding image from a collection of images given a caption. We follow the split of \citet{lee2018stacked} and use 1,000 images for validation and test each and train on the rest of the dataset. During fine-tuning, we follow UNITER \citep{chen2020uniter} and sample two negative text-image pairs along with a positive sample. We train for 5K steps with a batch size of 8 and a peak learning rate of $5\times10^{-5}$.

\paragraph{RefCOCO+} The referring expression task involves locating an image region given a natural language phrase. We follow ViLBERT \citep{lu2019vilbert} and conduct evaluation on the RefCOCO+ dataset. We use the bounding box proposals provided by \citet{yu2018mattnet}. For each box proposal, the model is trained to classify if it matches the reference phrase or not. A proposal box is considered correct if it has an IoU with the gold box larger than 0.5. We train for 12 epochs with a batch size of 32 and a peak learning rate of $5\times10^{-5}$.

\section{Data Accessibility}
The version of BookCorpus we used is downloaded from \url{https://github.com/jackroos/VL-BERT/blob/master/data/PREPARE_DATA.md}. The other datasets we used including Conceptual Captions, Open Images, VQA, \nlvr, Flickr30K, and RefCOCO+ are publicly available.

\end{document}